# A collaborative agent with two lightweight synergistic models for autonomous crystal materials research

Tongyu Shi[1,5*], Yutang Li[1,2,5*], Zhanyuan Li[1,5*], Qian Liu[1,3,5*], Jie Zhou[1,4], Wenhe Xu[1,2], Yang Li[1], Dawei Dai[4], Rui He[1,2], Wenhua Zhou[1,2], Jiahong Wang[1✉], Xue-Feng Yu[1,2✉]

1 Materials Artificial Intelligence Center, Shenzhen Institute of Advanced Technology, Chinese Academy of Sciences, Shenzhen, Guangdong, 518055, China

2 University of Chinese Academy of Sciences, Beijing 100049, China

3 Institute for Advanced Study, Chengdu University, Chengdu 610106, Sichuan, China.

4 Key Laboratory of Cyberspace Big Data Intelligent Security, Ministry of Education, Chongqing University of Posts and Telecommunications, 400065, Chongqing, China.

5 These authors contributed equally: Tongyu Shi, Yutang Li, Zhanyuan Li, Qian Liu

(Emails: jh.wang1@siat.ac.cn and xf.yu@siat.ac.cn)

## Abstract

Current large language models require hundreds of billions of parameters yet struggle with domain-specific reasoning and tool coordination in materials science. Here, we present MatBrain, a lightweight collaborative agent system with two synergistic models specialization for crystal materials research. MatBrain employs a dual-model architecture: Mat-R1 (30B parameters) as the analytical model providing expert-level domain reasoning, and Mat-T1 (14B parameters) as the executive model orchestrating tool-based actions. Entropy analysis confirms that this architecture resolves the conflict between tool planning and analytical reasoning by decoupling their distinct entropy dynamics. Enabled by this dual-model architecture and structural efficiency, MatBrain significantly outperforms larger general-purpose models while reducing the hardware deployment barrier by over 95%. MatBrain exhibits versatility across structure generation, property prediction, and synthesis planning tasks. Applied to catalyst design, MatBrain generated 30,000 candidate structures and identified 38 promising materials within 48 hours, achieving approximately 100-fold acceleration over traditional approaches. These results demonstrate the potential of lightweight collaborative intelligence for advancing materials research capabilities.

## 1. Introduction

The accelerated discovery of crystal materials is critical for energy conversion systems, electronic components, and aerospace structures.[1-3] Traditionally, material development has highly depended on iterative trial-and-error experimentation, resulting in cycles spanning 10-20 years.[4-8] Although advances in high-throughput screening and machine learning have successfully established vast databases and predictive models, the discovery process remains critically bottlenecked by the fragmentation of scientific workflows.[9-13] Specifically, the exploration of structure-property-synthesis relationships of material requires human experts to manually connect theoretical principles with specialized computational tools.[14-16] However, this manual method is inherently non-scalable and labor-intensive, severely limiting the exploration of vast chemical spaces. Therefore, autonomous systems capable of coupling domain-specific reasoning with tool orchestration are needed to accelerate the timeline from conceptualization to validation.

Large language models (LLMs) have demonstrated considerable potential for integrating knowledge, performing logical reasoning, processing information, and coordinating tasks across disciplines.[17-19] Recent advanced research, such as ChemCrow, has shown promising capabilities for tool-augmented chemical discovery.[20,21] However, applying general LLMs to materials science presents unique challenges.[22,23] General LLMs inherently lack grounding in rigorous 3D geometric constraints and quantum mechanical effects (e.g., lattice symmetries, band structures),[24-26] while also struggling with the rigid syntactic precision required for iterative, multi-step tool orchestration.[25,27-29] To mitigate these reasoning gaps, recent trends have favored scaling up to massive architectures (e.g., 600B+ parameters in DeepSeek-R1). Yet, such large-scale models involve substantial training overhead and deployment costs, restricting their accessibility to a few well-funded institutions and hindering real-time, interactive workflows.[28,30,31] More fundamentally, a critical optimization conflict arises from the distinct entropy dynamics required for these dual tasks. Specifically, reliable crystallographic analysis requires minimizing entropy to ensure deterministic accuracy, whereas robust tool orchestration requires a flexible, high-entropy policy to enable the adaptive planning and diverse reasoning exploration essential for navigating complex decision spaces.[32-34] Consequently, forcing a unified lightweight model to satisfy these conflicting objectives often leads to "entropy collapse", where the model either loses the flexibility to handle complex workflows or hallucinates deterministic facts.[35]

Herein, we introduce a lightweight collaborative agent with two synergistic models for autonomous crystal materials research (**Figure 1**a). MatBrain employs an architecture that decouples domain expertise and tool execution into two specialized models. Mat-R1 (30B parameters) specializes in crystallographic theory to provide expert-level interpretation and validity assessment for hypothesis generation and result analysis. To facilitate tool utilization, we developed an integrated crystal research toolkit following a standardized Model Context Protocol (named Mat-MCP). Mat-T1 (14B parameters) operates as the executive model, trained through reinforcement learning using MCP-formatted interaction data to master tool selection and workflow orchestration.The collaborative architecture enables seamless coordination through iterative interaction. Mat-T1 processes research queries and executes workflows, while Mat-R1 analyzes results and formulates scientific insights for subsequent investigations (Figure 1b).

Mechanistic analysis validates that this architecture functionally decouples the probabilistic exploration required for adaptive planning from the deterministic convergence essential for analytical reasoning. Comprehensive evaluations demonstrate that MatBrain substantially outperforms general-purpose LLMs, such as the 671B-parameter DeepSeek-R1, while reducing hardware barriers by over 95%, enabling high-performance inference on a single dual-NVIDIA 4090 workstation. Beyond these advantages, MatBrain exhibits practical utility across diverse materials research tasks, including crystal structure generation, property prediction, and synthesis pathway analysis. As a representative demonstration, we applied MatBrain to nitrogen fixation catalyst design, where the system generated 30,000 candidate structures and identified 38 promising materials through coordinated dual-model analysis, ultimately selecting $CoV_4S_8$ for experimental validation (Figure 1c). This complex high-throughput generation and screening process was completed within 48 hours, representing a ~100-fold acceleration compared to traditional research cycles that span months. These findings demonstrate a potential pathway toward accessible and autonomous materials research systems for accelerating materials discovery.

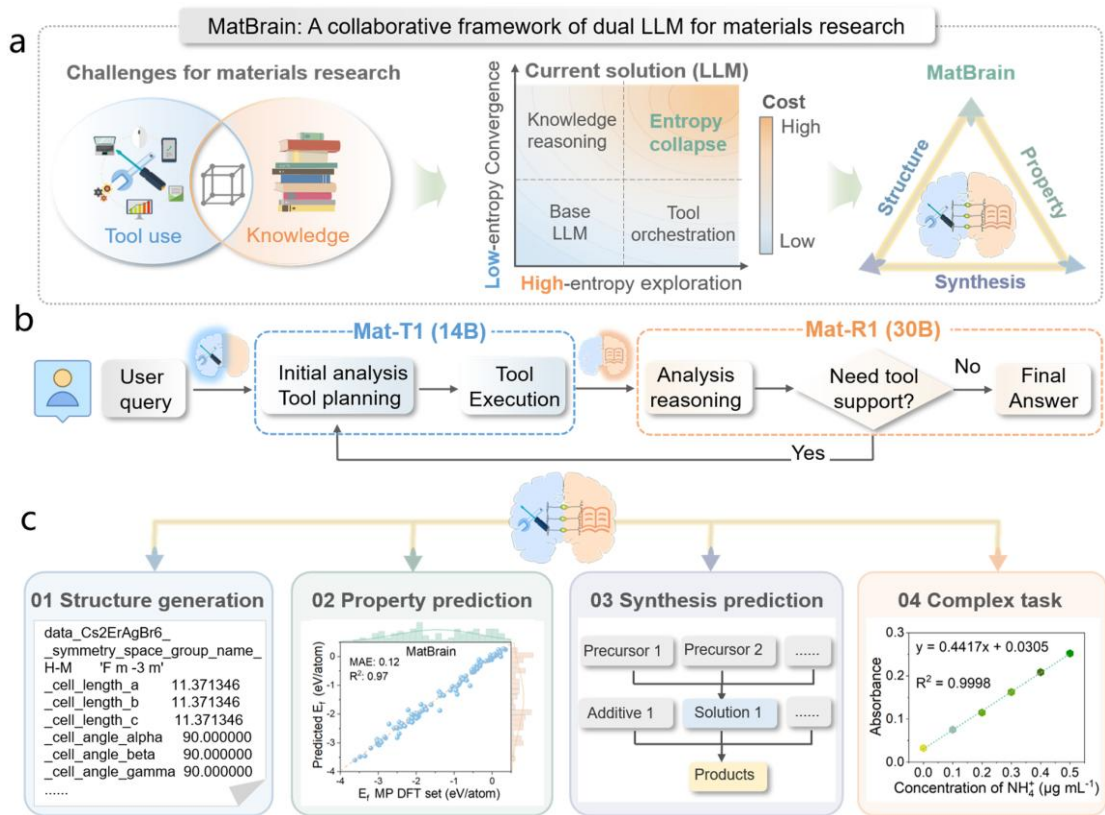

**Figure 1 | MatBrain: A lightweight dual-model collaborative framework for autonomous materials research.** (a) Overview of materials research challenges and solutions. Schematic illustrating the gap between tool use and knowledge (left), leading to entropy collapse in monolithic models due to conflicting entropy dynamics and

prohibitive computational costs (middle). MatBrain resolves this *via* a dual-model architecture unifying structure, property, and synthesis tasks (right). **(b)** Collaborative workflow between the two specialized lightweight models in MatBrain. **(c)** Representative applications demonstrating the versatility of MatBrain across materials research tasks: crystal structure generation, property prediction, synthesis pathway planning, and complex materials discovery tasks.

## 2. Results and discussion

### 2.1 Data curation and model training

The advancement of data-driven materials discovery depends critically on the quality and integration of the underlying knowledge base. While databases such as the Materials Project (MP) provide structured crystallographic data, essential insights regarding synthesis protocols and applications remain fragmented within unstructured scientific literature.[36,37] This separation has historically limited the development of unified frameworks capable of linking crystal structure and properties with synthesis pathways.[25] To bridge this gap, a standardized automated pipeline was established to integrate these heterogeneous sources (**Figure 2**a). By aligning structured database entries with literature-extracted text *via* unique identifiers (e.g., DOIs, MP-ids), a comprehensive corpus linking CIF data (structure information), JSON (property information), and PDF (literature data) was constructed (Figure S1). A generation-distillation strategy was then employed to convert this integrated data into the Mat-252K-SFT dataset (Figure S2), comprising 252,000 high-quality instruction-tuning pairs designed to ground the model in multidimensional materials knowledge (Figure S3, see more details in Methods).

Constructing a foundational model for materials science requires a rigorous balance between intrinsic physical-chemical knowledge and reasoning capacity.[38] To identify optimal architectures, initial language modeling loss on the training corpus and performance on scientific benchmarks (GPQA[39], HLE[40]) were utilized as proxies for learning potential (Figure 2b). This comparative analysis highlighted Qwen3-30B-A3B and Qwen3-14B as offering the most favorable trade-off between computational efficiency and scientific capacity. Consequently, Qwen3-30B-A3B was selected as the foundation model for the analytical model (Mat-R1) due to its superior reasoning capacity, while the lighter Qwen3-14B was chosen for the executive model (Mat-T1) to facilitate resource-efficient reinforcement learning.

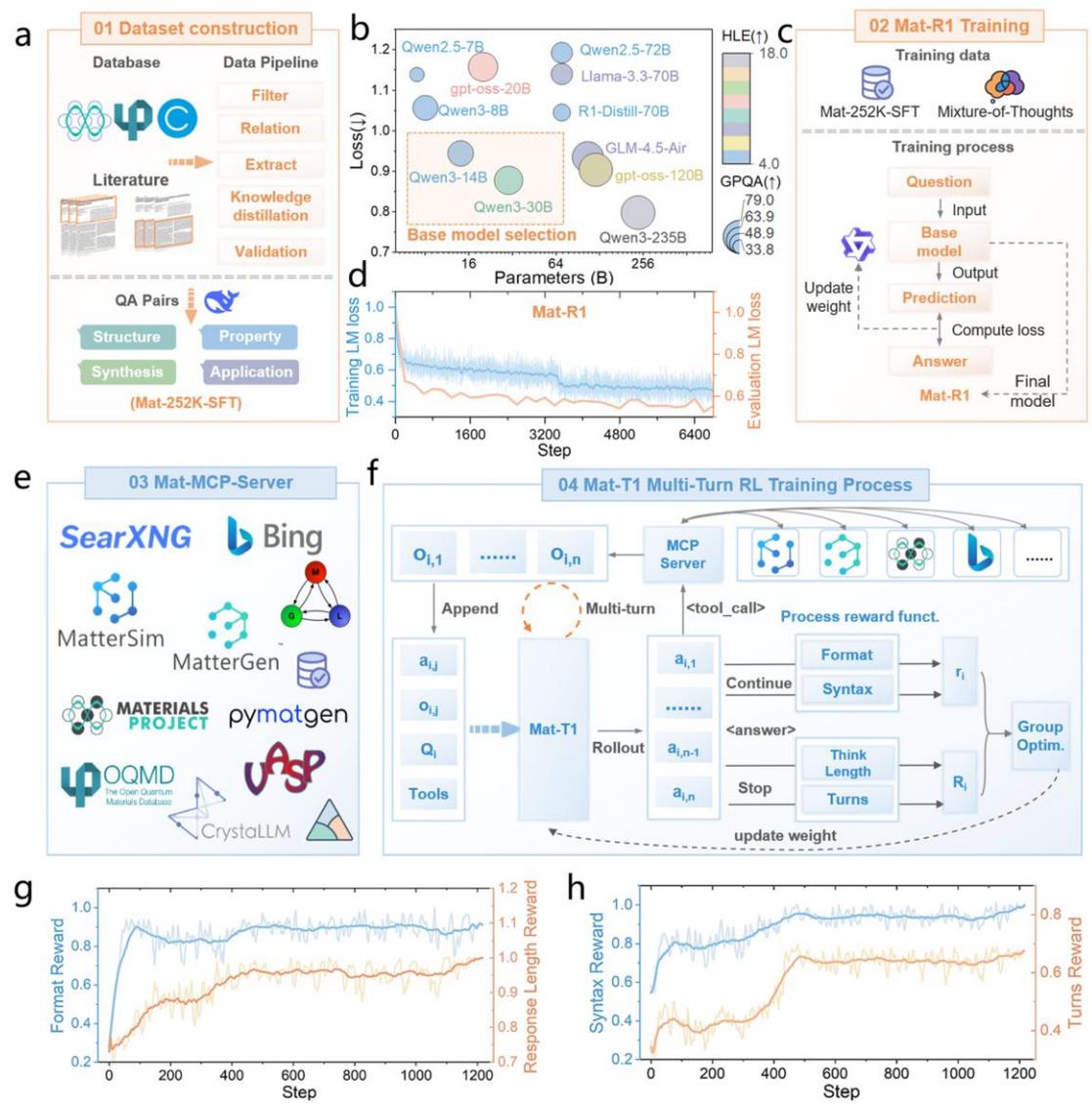

**Figure 2 | Construction and training paradigm of the MatBrain dual-model system.** **(a)** Automated data curation pipeline. Heterogeneous data from crystallographic databases and unstructured scientific literature are integrated to construct the Mat-252K-SFT dataset. **(b)** Base model selection strategy. The dashed box highlights the selection of Qwen3-30B and Qwen3-14B as optimal foundations for the analytical and executive models, respectively. **(c)** Training workflow of Mat-R1. The analytical model is trained on the domain-specific Mat-252K-SFT dataset augmented with open-

source scientific reasoning trajectories (open-r1/Mixture-of-Thoughts) to enhance domain expertise and logical capabilities. **(d)** Training results of Mat-R1. The convergence in both training and evaluation language modeling (LM) losses indicates the successful acquisition of materials science knowledge. **(e)** Architecture of the Mat-MCP ecosystem. A standardized MCP server encapsulates diverse computational tools, enabling autonomous tool execution *via* a unified interface. **(f)** Multi-turn reinforcement learning framework for Mat-T1. **(g)** Training curves of the format reward and response length reward. **(h)** Training curves of the syntax reward and the turns reward.

As shown in Figure 2c, Mat-R1 was developed *via* full-parameter supervised fine-tuning (SFT).[41] The training corpus combined the specialized Mat-252K-SFT dataset with a broad scientific reasoning subset (open-r1/Mixture of Thought), a strategy designed to deepen crystal structure-property-synthesis-application comprehension while preserving general logical acuity. To reinforce domain specialization, the system prompt was customized and self-identity data were utilized to align the interaction style with that of an expert materials scientist. The effectiveness of this training strategy is demonstrated in Figure 2d, where training and validation losses declined from approximately 1.0 to 0.5, indicating a significant improvement in domain-specific capabilities.[42]

To complement theoretical reasoning with precise execution, Mat-T1 was engineered as an executive model optimized for autonomous tool orchestration. As presented in Figure 2e, the Mat-MCP platform was developed following the Model Context Protocol (MCP) to standardize and integrate the fragmented ecosystem of materials science tools.[45] This modular system integrates essential functions, including information retrieval, structure generation, relaxation, and property prediction into standardized Docker containers to ensure stability and scalability. While standard agent frameworks (e.g., ReAct) frequently fail during complex, multi-step scientific workflows, Mat-T1 was trained *via* reinforcement learning (RL) using the Decoupled Clip and Dynamic sAmpling Policy Optimization (DAPO) algorithm (Figure 2f).[46-49]

A fundamental challenge in the training of scientific models is presented by the non-uniqueness of solution trajectories in materials design, where rigid outcome-based supervision often proves sparse or misleading. To address this, the deterministic nature of scientific tools is utilized as an intrinsic validator, where valid execution serves as a proxy for adherence to physical constraints.[50,51] Consequently, rather than relying solely on final answer matching, we reformulated the optimization objective to prioritize process fidelity and executability. Through this strategy, a self-supervised framework is established to mitigate "reward hacking", ensuring that accuracy is derived from rigorous tool execution rather than answer memorization.[52] Accordingly, policy optimization utilizes a composite reward function that treats syntactic validity, format compliance, and workflow efficiency as the target outcomes.

Mat-20K-RL was developed as a dataset of 20,000 open-ended queries devoid of ground-truth labels for model training. As illustrated by the training dynamics in Figures 2g-2h, the format reward converged rapidly within the initial 200 steps, indicating the efficient acquisition of the iterative reasoning-acting-observing loop. Subsequently, the evolution from rudimentary tool utilization to diverse tool selection across multiple turns is reflected by the steady ascent and convergence of rewards associated with interaction turns, response length, and syntax. This cognitive deepening is further evidenced by a substantial increase in the average total sequence length (input and output) from approximately 12,000 to 19,000 tokens (Figure S4).[53] Crucially, the Actor entropy of Mat-T1 exhibits a distinct upward trend (Figure S4), contrasting with the mode collapse often observed in standard reinforcement learning. Attributable to an unconstrained policy space (KL coefficient = 0.0), this trajectory signifies the active exploration of diverse tool utilization paths rather than a convergence to deterministic shortcuts.[33,35] Concurrently, the Critic reward steadily climbed to ~0.95 alongside zero-centered advantage oscillations, indicating robust training stability. Demonstrating this practical capability, Figure S5 shows how Mat-T1 accurately analyzes user intent, orchestrates tool sequences, and conducts workflows to deliver validated scientific outcomes.

**2.2 Dual-model collaborative architecture and evaluation**

To systematically validate the efficacy of the MatBrain architecture, a comprehensive benchmark covering generation, classification, and regression tasks was established using the curated Mat-252K test set. First, the independent representational capacity of the analytical model (Mat-R1) was assessed against leading general-purpose foundational models, including DeepSeek-R1, GPT-5, and Gemini 2.5 Pro. As shown in **Figure 3**a, while general models demonstrate competence in classification tasks (e.g., is_magnetic), they exhibit near-total failure in tasks governed by rigorous 3D geometric logic, specifically structure similarity (sim_structure) and CIF syntax (cif_syntax), with scores approaching zero. In contrast, Mat-R1 demonstrates substantial performance gains across all metrics, a direct consequence of full-parameter fine-tuning. Notably, even when general models are augmented with ReAct-based agents, their ability to generate valid structures remains significantly inferior to the purely knowledge-driven Mat-R1 (Figure S6). This demonstrates the critical importance of targeted domain knowledge training, which enables Mat-R1 to establish a deep semantic mapping between natural language and crystal geometry (CIF). Furthermore, scaling analysis indicates that the 30B parameter scale represents an optimal balance between reasoning performance and computational efficiency (Figure S7 and Table S1).

The contribution of the executive model (Mat-T1) was subsequently isolated to evaluate the impact of specialized tool learning. Performance was compared across three distinct configurations: a baseline model without tool access, a model augmented *via* the standard ReAct strategy, and the fully integrated Mat-T1 system. Figure 3b and Figure S8 reveal a distinct hierarchy in capability. Although standard ReAct strategies help mitigate hallucinations in basic tasks like stability prediction and yield a 10-20% improvement, these methods fail in complex scenarios that require precise parameter configuration, such as magnetic ordering and structure similarity analysis. In contrast, the integration of Mat-T1 yields a significant performance increase over ReAct methods. This disparity highlights a critical limitation of prompt engineering in scientific domains: complex tools possess tacit knowledge that cannot be fully conveyed *via* context prompts alone. By internalizing this procedural knowledge through reinforcement learning, Mat-T1 outperforms general LLMs and enables the precise orchestration of complex scientific workflows.

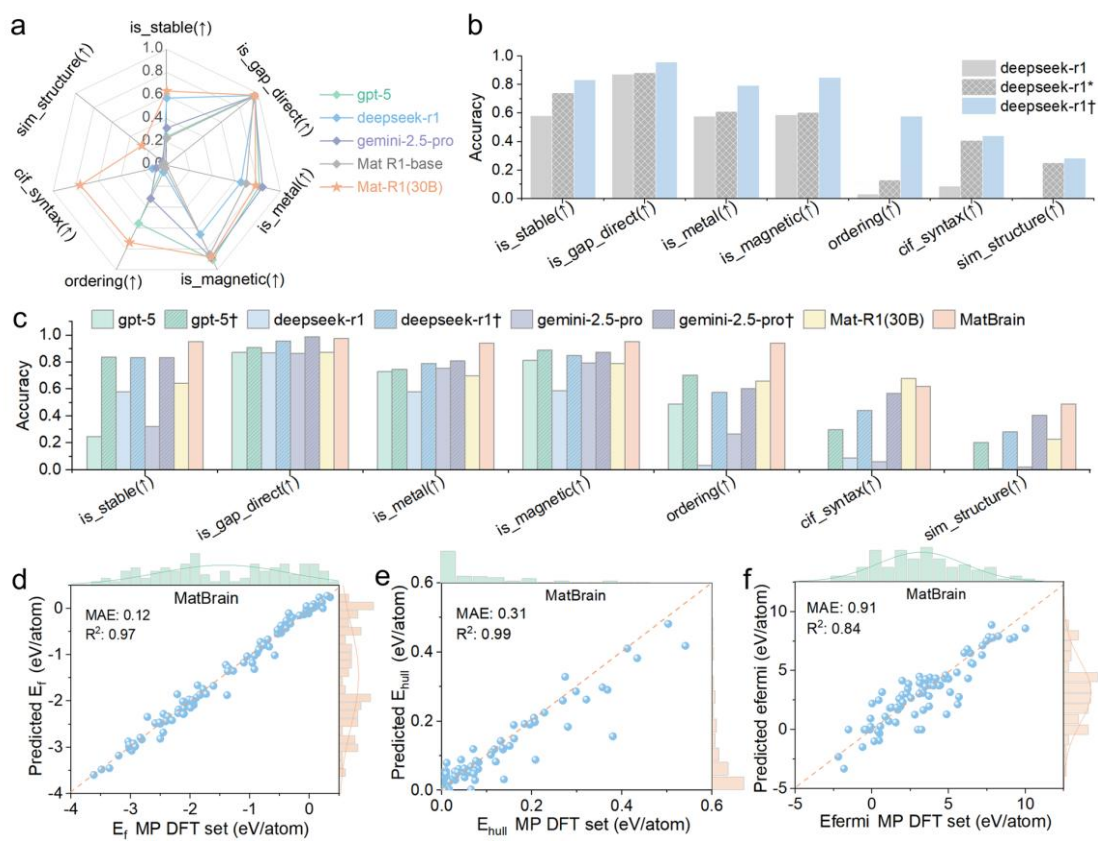

**Figure 3 | Systematic evaluation and performance benchmarking of the MatBrain system.** (a) Capability comparison of different models. Mat-R1 (orange star) exhibits balanced and robust performance, significantly outperforming general LLMs in domain-specific metrics. (b) Ablation study on tool-utilization strategies using DeepSeek-R1 as the base model. The comparison includes the base model, the model with standard ReAct strategy (*), and the model integrated with Mat-T1 (†). (c) Comprehensive accuracy comparison between MatBrain, Mat-R1 (30B), and state-of-the-art general LLMs (GPT-5, DeepSeek-R1, Gemini-2.5-Pro), evaluated both in their default states and with Mat-T1 integration (†), across seven key materials science tasks. (d-f) Comparison of MatBrain-predicted energetic properties against ground truths from the MP database. The plots display results for (d) formation energy ($E_f$), (e) energy above hull ($E_{hull}$), and (f) Fermi energy ($E_{fermi}$). The high coefficients of determination ($R^2$) and low Mean Absolute Errors (MAE) demonstrate the precision of MatBrain in quantitative regression tasks through synergistic knowledge reasoning and tool execution.

Building upon the specialized training of the analytical and executive models, the MatBrain system implements a synergistic dual-model architecture (Figure 1b). Mat-T1 specializes in procedural knowledge, responsible for translating abstract scientific inquiries into precise sequences of Mat-MCP tool invocations while ensuring the syntactical validity of execution. Complementing this, Mat-R1 focusing on domain knowledge to provide theoretical analysis and interpret the computational results generated by Mat-T1. Notably, this collaboration proceeds through a dynamic "generation-verification-feedback" loop rather than a linear sequence. As illustrated in Figure S9, Mat-R1 evaluates information completeness at each step, autonomously triggering iterative tool deployment by Mat-T1 when necessary. This adaptive mechanism enables the system to self-correct and robustly resolve intricate, multi-step materials analysis workflows (Figure S10). Significantly, this decoupled design breaks down the barrier for deployment. As shown in Table S2, MatBrain reduces infrastructure costs by over 95%, transitioning requirements from expensive H100 clusters (~$600K) to standard laboratory workstations (~$15K), thereby making state-of-the-art material analysis immediately accessible to the broader scientific community.

The performance of this lightweight collaborative system was further evaluated. As demonstrated in Figure 3c, the fully integrated MatBrain system achieves the highest accuracy across all seven tasks, systematically outperforming both the standalone Mat-R1 and tool-augmented general LLMs (including GPT-5 and DeepSeek-R1). Beyond classification and generation, performance on quantitative regression tasks was evaluated using formation energy ($E_f$), hull energy ($E_{hull}$), and Fermi level ($E_{fermi}$). As illustrated in Figures 3d–f, predictions from MatBrain tightly converge along the diagonal, achieving coefficients of determination ($R^2$) to 0.97, 0.99, and 0.84, respectively. Correspondingly, Mean Squared Errors (MSE) are orders of magnitude lower than those of general baselines (Tables S3–S5). In sharp contrast, general models exhibit catastrophic failure in regression tasks (Figures S11–S13), producing random scatter or mean-value guessing. Fundamentally, this performance gap stems from a paradigm shift in the solution pathway: from direct parameter-based estimation to tool-mediated derivation. General LLMs attempt to regress physical quantities by treating them as textual tokens to be predicted directly from internal training distributions, a process that relies on memorization and inherently lacks physical constraints. MatBrain thus achieves superior performance over general super-sized models, offering a viable pathway for low-cost, locally deployable "Green AI for Science".

**2.3 Dual-model decoupling from a Shannon entropy perspective**

Distinct from the cross-entropy utilized in model training, Shannon entropy serves as a classic metric for measuring information uncertainty, intuitively revealing the intrinsic cognitive state of the model at the token level during inference.[54] To investigate the differences between the cognitive modes of Mat-T1 and Mat-R1 at a mechanistic level, we conducted a comparative analysis of token-level Shannon entropy trajectories before and after training.[35] **Figure 4** visualizes the entropy evolution of different models given the same input, while also providing a statistical view of the feature distributions. First, the base models show distinct entropy differences when addressing identical tasks. As illustrated in Figures 4a–b, Qwen3-30B-A3B demonstrates a natural convergence trend (mean = 0.673) when directly answering scientific questions, which aligns with the knowledge stability required for the analytical model. Conversely, when Qwen3-14B uses tools *via* Mat-MCP, it displays higher average entropy (mean = 0.878) with significant fluctuations, highlighting the exploratory nature essential for the executive model.

The training process further amplifies these differences. After SFT, Mat-R1 exhibits a deterministic low-entropy profile (Figure 4c), where the average entropy decreases significantly from 0.673 to 0.483. This indicates that Mat-R1 has learned a deterministic distribution of crystallographic knowledge, allowing it to suppress errors and output answers with higher confidence.[55] In contrast, Mat-T1 develops a high-entropy exploratory pattern after RL optimization (Figure 4d), where the average entropy does not decrease but rather increases from 0.878 to 0.974. This increase demonstrates that Mat-T1 learns to expand the search space and sample diverse tool paths to acquire authoritative external data, rather than converging prematurely to local suboptimal solutions.[53]

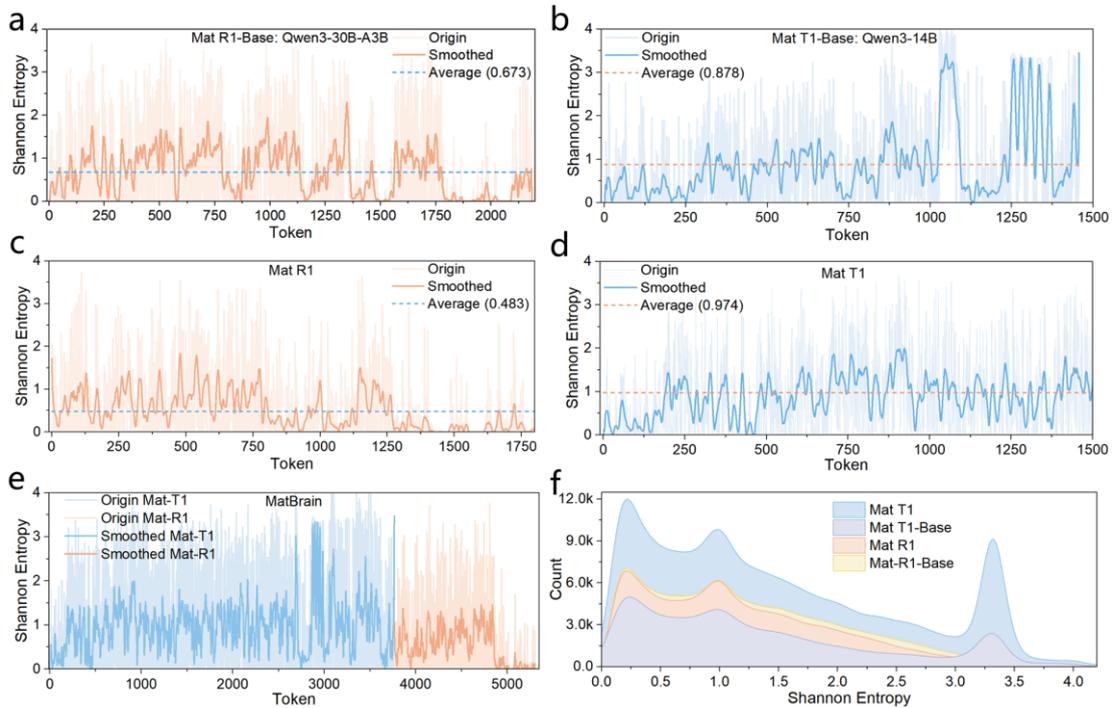

**Figure 4 | Entropic differentiation and statistical validation of dual-model cognitive modes. (a)** Entropy trajectory of the analytic base model (Qwen3-30B-A3B) on a scientific task, showing a natural convergence trend (mean = 0.673). **(b)** Entropy trajectory of the executive base model (Qwen3-14B) utilizing Mat-MCP tools, displaying an elevated average entropy (mean = 0.878) with significant fluctuations. **(c)** Entropy evolution of Mat-R1 following SFT. The model exhibits a deterministic, low-entropy profile (mean = 0.483), indicating high confidence in crystallographic reasoning. **(d)** Entropy evolution of Mat-T1 following RL optimization. The model develops a high-entropy exploratory pattern (mean = 0.974), reflecting adaptive planning over a broadened search space. **(e)** Collaborative entropy dynamics within MatBrain. The injection of high-entropy observations from Mat-T1 facilitates rapid convergence in the reasoning phase of Mat-R1. **(f)** Kernel density estimation of entropy distributions across 200 test samples. Mat-T1 features a distinct entropy peak at ~3.3 bits (aligning with the effective tool action space), whereas the distribution of Mat-R1 shows monotonic decay.

Figure 4e unveils the effectiveness of the dual-model collaborative mechanism. When Mat-T1 provides observations acquired through high-entropy exploration as context, the inference process of Mat-R1 shows stronger convergence. Although the overall average entropy of the output from Mat-R1 increases slightly to 0.533 (reflecting the processing of complex information), the average entropy of the core conclusion section (<answer>) drops precipitously from 0.12 to 0.05. This result confirms that the collaborative mechanism successfully converts the exploration breadth of Mat-T1 (high entropy) into the decision precision of Mat-R1 (low entropy), achieving an effective transformation from information gain to conclusion confidence. Furthermore, we randomly sampled 200 question-answer pairs from the test set, calculated their token-level entropy values across four models, and plotted kernel density estimation (KDE) curves. As shown in Figure 4f, Mat-T1 exhibits a distinctive peak emerging in the high-entropy interval (~3.3 bits), a feature completely absent in the distributions of Mat-R1 and R1-Base. This peak quantifies the decision uncertainty space necessary for agents when navigating complex tool planning and multi-branch invocations. In contrast, the distribution of Mat-R1 presents monotonic decay, consistent with the deterministic characteristics of model reasoning. This statistically significant separation validates the necessity of functionally decoupling adaptive planning and analytical reasoning in the dual-model architecture from an information-theoretic perspective.

**2.4 Practical application of representative research**

To demonstrate the capacity of MatBrain for end-to-end research autonomy, the system was deployed across six representative case studies covering the complete materials research lifecycle, from structure generation to experimental design (**Figure 5**). The workflow begins with de novo structural determination, where MatBrain autonomously constructed the $Cs_2ErAgBr_6$ lattice through a multi-step search-and-generation protocol and effectively mitigated the hallucination risks common in pure language models (Figure 5a). Its capability is particularly evident when handling complex disordered structures like $CsPb(Cl_{0.2}Br_{0.4}I_{0.4})_3$ (Figure 5b). While conventional generative models failed to process the fractional occupancies, MatBrain implements fractional occupancy modeling and Vegard's Law refinement.[56] Such adaptability stems from combining reasoning with tool execution, guaranteeing the crystallographic validity of the generated structures even for complex, non-stoichiometric cases.

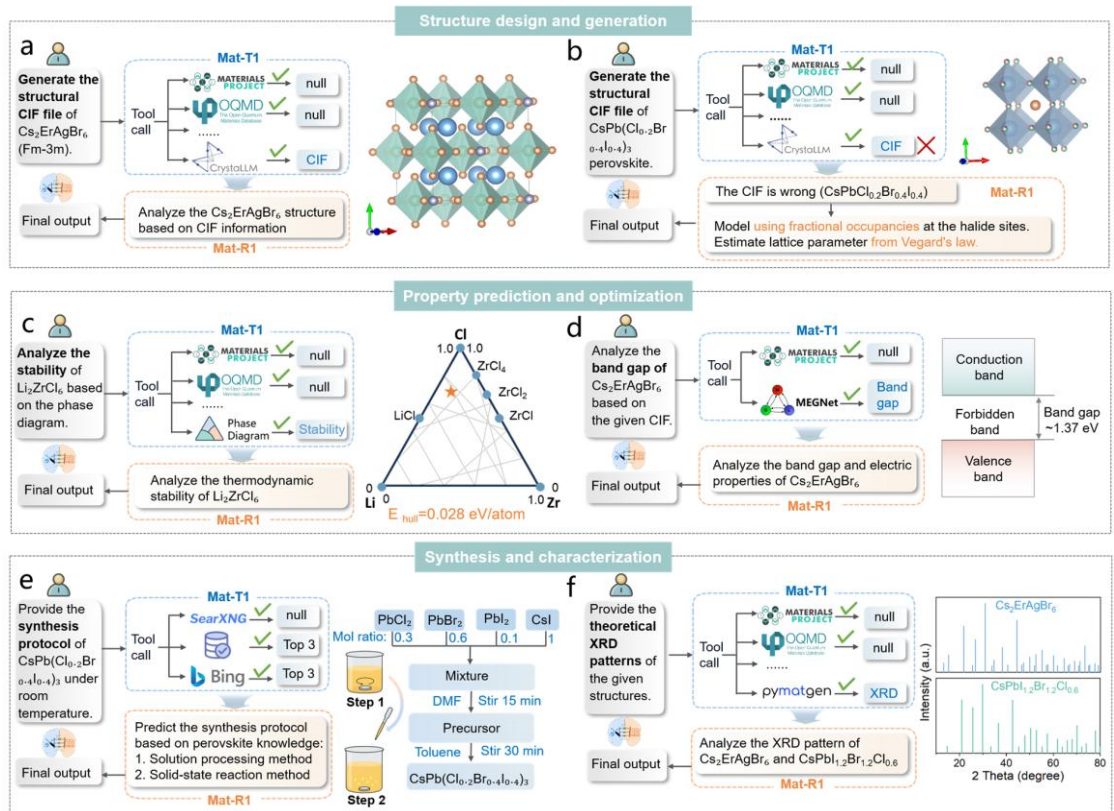

**Figure 5 | Practical application of MatBrain across the materials research lifecycle.** **(a)** Autonomous structure generation for the rare-earth double perovskite $Cs_2ErAgBr_6$. The system generates a valid CIF structure after confirming the absence of existing database records. **(b)** Expert-level modeling of the complex mixed-halide perovskite $CsPb(Cl_{0.2}Br_{0.4}I_{0.4})_3$. MatBrain successfully constructs the non-stoichiometric lattice using fractional occupancy and Vegard's Law. **(c)** Thermodynamic stability analysis of $Li_2ZrCl_6$. The system autonomously constructs the Li-Zr-Cl ternary phase diagram, identifying the compound as a metastable phase ($E_{hull}$=0.028 eV/atom). **(d)** Electronic property prediction for $Cs_2ErAgBr_6$. The calculated bandgap of 1.37 eV indicates strong potential for visible-light optoelectronic applications. **(e)** Intelligent design of a synthesis protocol. MatBrain formulates a detailed room-temperature synthesis pathway for the mixed-halide perovskite, including precursor selection and reaction conditions. **(f)** Theoretical characterization fingerprints. Simulated XRD patterns for $Cs_2ErAgBr_6$ provide reference standards for experimental validation.

Following the structural definition, the system demonstrated versatility by tailoring the analysis to the specific needs of different candidates. For the electrolyte $Li_2ZrCl_6$ (Figure 5c), the priority was thermodynamic stability. To assess this, MatBrain integrated a specialized phase stability tool combining CHGNet calculations (for predicting accurate formation energies), MP database queries, and pymatgen phase diagram construction utilities. By invoking this PhaseDiagram tool, MatBrain successfully constructed a ternary Li-Zr-Cl phase diagram. Interpreting the resulting topology, the system identified $Li_2ZrCl_6$ as a metastable phase ($E_{hull}$=0.028 eV/atom). MatBrain further reasoned that while the material is kinetically stable at room temperature, precise thermal management is required to prevent decomposition into stable sub-phases. Concurrently, for the optoelectronic candidate $Cs_2ErAgBr_6$ (Figure 5d), the analytical focus shifted to functional utility. Here, the system employed the MEGNet model to predict electronic structure, calculating a bandgap of 1.37 eV. It then evaluated the overlap of this bandgap with the solar spectrum, analytically confirming the potential of the material for efficient visible-light harvesting.

Crucially, theoretical predictions must eventually be translated into experimental reality. To bridge this divide, MatBrain formulated a precise, step-by-step synthesis protocol for the complex perovskite, deriving precursor stoichiometries and reaction conditions from chemical principles rather than simple text retrieval (Figure 5e). To close the verification loop, it simulated the theoretical X-ray diffraction (XRD) fingerprints for the designed materials (Figure 5f), providing researchers with definitive reference standards for product validation. Collectively, these case studies confirm that MatBrain is not merely a calculation engine, but a comprehensive research agent capable of driving materials innovation from initial concept to experimental validation.

**2.5 Autonomous discovery and validation of nitrogen fixation catalysts**

Having demonstrated the versatility of the system in discrete tasks, we further integrated MatBrain into a complex, real-world research campaign to validate its capacity for end-to-end discovery. **Figure 6** illustrates the utilization of MatBrain for the design of nitrogen fixation catalysts, spanning from conceptual ideation to candidate identification. The study commenced with a natural nitrogenase mechanism (Figure

6a), and a workflow was initiated to face the open-ended challenge: Mat-T1 retrieved relevant knowledge on nitrogenase cofactors, while Mat-R1 integrated this information with its internal knowledge. This synergistic analysis identified that vanadium-based nitrogenases are underutilized in artificial catalysis compared to their molybdenum counterparts, leading the system to propose ternary M-V-S (M=Fe, Co, Ni) metal sulfides as a strategic design direction.

Subsequent to this hypothesis, the system was assigned the task of exploring the identified chemical space by researchers. Mat-T1 orchestrated the parallel generation of 30,000 initial structures for the three target systems *via* the MatterGen tool. To process this massive candidate pool efficiently, a systematic seven-step high-throughput screening workflow was established to progressively refine candidates across chemical, structural, and thermodynamic criteria (Figure 6b). The pipeline began by strictly enforcing the ternary Fe/Co/Ni-V-S composition, removing ~10% of the pool containing impurities or deviations. Subsequently, Mat-T1 applied chemical rules to filter for valence plausibility, rejecting unstable oxidation states (e.g., $V_2Fe_6S_4$) and narrowing the field to 21,954 entries. Structural redundancy was then addressed through symmetry analysis and structure matching. Utilizing tolerance parameters optimized by Mat-R1 (site tolerance 0.3, lattice tolerance 0.2, angle tolerance 5 degrees), MatBrain retained a diverse range of unique structures without expending computational resources on duplicates. A geometric validity check further filtered the pool to 10,128 viable candidates by excluding physically impossible atomic overlaps.

The most critical filter was thermodynamic stability, the primary determinant of synthetic feasibility. To address the computational challenge of assessing thousands of structures, the system deployed a specialized phase diagram tool integrated within the Mat-MCP framework, which automatically computes the energy above hull for each candidate relative to the stable convex hull. Mat-R1 imposed a rigorous thermodynamic threshold of $\leqslant 0.025$ eV/atom, a standard chosen to maximize experimental realizability. This stringent filtration proved highly selective, eliminating over 99.6% of the structures and distilling the pool to 42 highly stable candidates. Subsequently, electronic properties were evaluated. The MEGNet calculations indicated that all 42 candidates are semiconductors with bandgap smaller than 2.0 eV, which is considered optimal for electrocatalytic electron transfer as identified by Mat-R1. A final novelty check against existing databases revealed that 38 of these structures (90.5%) had not been reported previously.

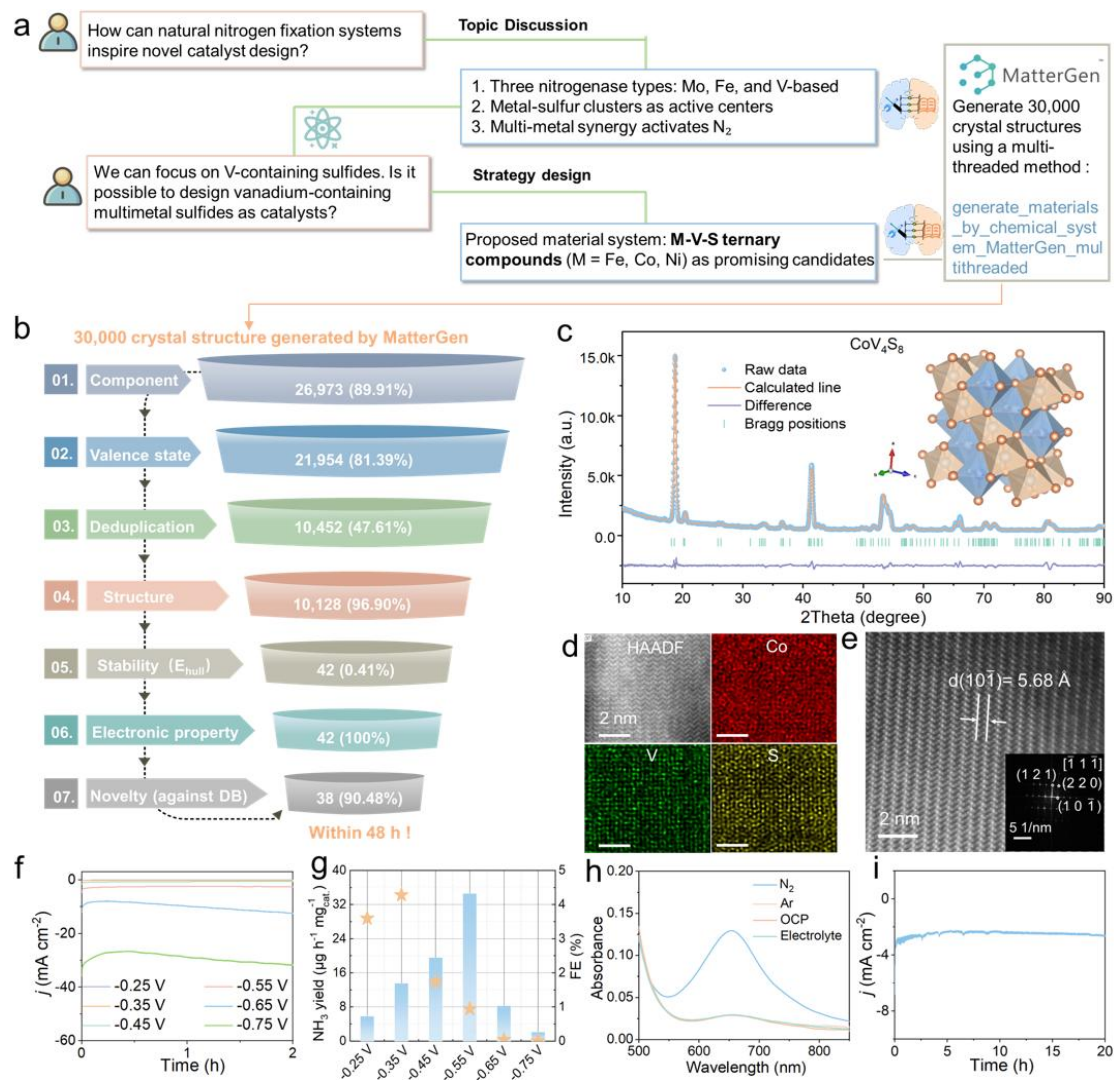

**Figure 6 | End-to-end autonomous discovery of bio-inspired nitrogen fixation catalysts. (a)** Interactive design process from nitrogenase analysis to M-V-S catalyst proposal and 30,000 structure generation. **(b)** A seven-step screening funnel reduces 30,000 candidates to 38 novel materials through sequential filters. **(c)** XRD Rietveld refinement confirming the C2/m space group of $CoV_4S_8$. **(d)** Energy-dispersive X-ray spectroscopy mapping shows the uniform distribution of Co, V, and S elements. **(e)** Scanning transmission electron microscopy images reveal the pristine crystal lattice and atomic arrangement. **(f)** Chronoamperometry curves for $CoV_4S_8$ NS at various applied

potentials. **(g)** NH$_3$ yields and Faradaic efficiencies for CoV$_4$S$_8$ NS at various applied potentials. **(h)** UV-Vis spectra of the electrolytes stained with the indophenol reagent after 2 h of electrolysis under different test conditions. **(i)** Long-term NRR electrolysis for 20 h on CoV$_4$S$_8$ NS electrode at −0.55 V.

Notably, the MatBrain architecture autonomously executed this entire screening workflow within just 48 hours. Quantitatively, this represents an approximately 100-fold acceleration over traditional human-expert workflows, demonstrating a significant efficiency improvement over traditional research cycles that span months. This capability highlights the distinct advantage of MatBrain in lowering the technical barrier for non-specialists while upholding strict scientific rigor, ultimately accelerating materials innovation.

From the final shortlist, CoV$_4$S$_8$ was selected by the researcher for experimental validation, considering the high vanadium content (V: Co = 4:1) and robust stability. Following the synthesis protocol generated by MatBrain, the target material was successfully prepared. As shown in Figure 6c and Figure S14, XRD and structure refinement analysis confirmed the C2/m space group. Chemical analysis verified the precise stoichiometric ratio of Co:V:S, ensuring phase purity (Figure S15). Morphological characterization *via* scanning electron microscopy (SEM) revealed a layered structure, which facilitated exfoliation into nanosheets (Figure S16). Elemental mapping demonstrated a uniform distribution of all constituent elements (Figure 6d), while atomic-resolution imaging revealed a shape crystal lattice (Figure 6e and Figure S17). These results provide direct experimental validation of the computational design, confirming that the system successfully guided the research from a theoretical concept to a realizable material.

To validate the functional utility of the designed material, the nitrogen reduction reaction (NRR) performance of CoV$_4$S$_8$ nanosheets (NS) was evaluated in a proton exchange membrane H-cell. Ammonia (NH$_3$) and hydrazine (N$_2$H$_4$) were quantified *via* colorimetric UV-Vis spectroscopy, calibrated against standard curves (Figures S18-S19). Comparative Linear Sweep Voltammetry (LSV) revealed a distinct current density enhancement in $N_2$-saturated electrolyte compared to Ar-saturated controls across the −0.30 V to −0.68 V range (Figure S20), providing initial confirmation of NRR activity. Subsequent chronoamperometry tests conducted from −0.25 to −0.75 V displayed stable current responses (Figure 6f), accompanied by a potential-dependent evolution in catholyte color and UV absorption intensity (Figure S21).

As shown in Figure 6g, quantitative analysis revealed a volcano-type performance trend. A maximum $NH_3$ yield of 34.6 $\mu g \cdot h^{-1} \cdot mg_{cat.}^{-1}$ was achieved at −0.55 V, while the peak Faradaic efficiency (FE) of 4.3% was observed at −0.35 V. At more negative potentials, both metrics declined significantly, attributed to the intensified competition from the hydrogen evolution reaction (HER). Crucially, no hydrazine byproduct was detected at any potential (Figure S22), indicating high product selectivity. To rigorously verify the nitrogen source, control experiments were performed under argon atmosphere and open-circuit conditions, yielding negligible ammonia signals (Figure 6h and Figure S23). Detailed impurity analysis confirmed that background contamination accounted for less than 3% of the total yield (Figures S24-25), validating that the detected ammonia originates from electrochemical $N_2$ reduction. Finally, the catalyst demonstrated robust durability, maintaining stable yields over 5 consecutive cycles (Figure S26) and sustaining steady current density during a 20-hour long-term test (Figure 6i). These results confirm $CoV_4S_8$ as a stable and effective NRR catalyst, validating the translational capability of the MatBrain discovery pipeline.

## 3. Discussion

In this work, we presented MatBrain, a lightweight dual-model collaborative agent that addresses the critical challenges of applying large language models to crystal materials science. By coordinating the analytical capabilities of Mat-R1 with the executive proficiency of Mat-T1, the system achieves expert-level performance while operating on lightweight architectures. Through quantitative benchmarking on standard regression and classification tasks, MatBrain demonstrates superior performance compared to state-of-the-art general models. Beyond benchmarks, the system exhibits broad applicability across four distinct research domains: crystal structure generation, property prediction, synthesis pathway analysis, and characterization assistance. This comprehensive capability was exemplified by a complex and high-throughput discovery workflow for nitrogen fixation catalysts. In this representative case study, MatBrain autonomously generated 30,000 candidate structures and executed a multi-stage screening process to identify 38 promising candidates. This workflow was completed within 48 hours, representing a 100-fold acceleration compared to traditional methods and significantly lowering the technical barrier for non-specialists.

The significance of this research lies in three key contributions. First, we established Mat-MCP, a standardized material research tool ecosystem. By integrating tools for structure prediction, calculation, and synthesis analysis into a unified, containerized infrastructure, Mat-MCP resolves the fragmentation of computational tools and provides a robust foundation for future AI-driven materials research. Second, the success of the dual-model architecture validates a crucial methodological paradigm: effectively decoupling knowledge reasoning from tool execution enables smaller, specialized models to solve complex scientific problems more efficiently than massive general-purpose models. Third, regarding accessibility, MatBrain reduces the hardware deployment barrier by over 95%, enabling high-performance inference on single workstations. This significantly lowers both deployment and operational computational costs, making advanced AI capabilities accessible to individual laboratories.

The system is currently limited by the quality of available crystallographic databases and the inherent accuracy of standard DFT calculators. Looking forward, the modular design of the Mat-MCP protocol ensures high extensibility, allowing for the seamless integration of more tools. Besides, future iterations will focus on integrating multi-modal data, including microscopy images and spectral data, to enhance the perception capabilities of the analytical model. Ultimately, we envision integrating MatBrain with robotic automation platforms to close the loop between computational prediction and physical synthesis, thereby realizing fully autonomous and self-driving materials laboratories.

## 4. Method：

### 4.1 Construction of the Mat-252K-SFT dataset

#### 4.1.1 Data collection and curation

The training corpus was primarily curated from two sources: (1) peer-reviewed academic literature; and (2) materials databases, specifically the Materials Project (MP), the Open Quantum Materials Database (OQMD) and the Inorganic Crystal Structure Database (ICSD). First, crystallographic information files (CIFs) and corresponding computed property metadata (JSON) were systematically retrieved from these public databases. Subsequently, utilizing the Digital Object Identifiers (DOIs) associated with each database entry, the original research articles were acquired in PDF format *via* institutional library access. To guarantee data integrity, automated validation protocols were implemented to verify the completeness of the downloaded PDF files, and any corrupted or unreadable documents were rigorously excluded. Finally, a high-fidelity foundational database was established by creating precise one-to-one mappings between structures (CIF), properties (JSON), and literature (PDF) using unique identifiers such as MP-ids, ICSD-ids, and DOIs.

#### 4.1.2 Data preprocessing

To process the unstructured scientific literature, we utilized MinerU to convert raw PDF documents into a standardized, LLM-compatible Markdown format.[57] These documents were subsequently segmented into discrete text chunks based on structural delimiters. To ensure robust semantic categorization, we implemented a multi-model ensemble strategy. Each text chunk was processed by three state-of-the-art LLMs (DeepSeek-V3, GPT-4o, and Claude-3.5), and a majority voting mechanism was employed to classify the segments into one of four distinct domains: Structure, Properties, Synthesis, or Applications. This ensemble approach effectively mitigates the classification bias inherent in single-model predictions. Leveraging this classification, we deployed LangExtract to precisely isolate and extract specific text passages of synthesis pathways and application scenarios.[58] The resulting collection of refined text content served as the foundational corpus for constructing the Mat-252K-SFT dataset.

**4.1.3 Dataset construction *via* the generate-distill strategy**

The Mat-252K-SFT dataset was constructed using a rigorous generate-distill framework, orchestrated through three sequential phases:

**Context-aware question generation:** We utilized DeepSeek-V3 to formulate domain-specific scientific inquiries. By feeding the refined text chunks as context, combined with meticulously engineered prompts, the model was directed to generate high-quality questions encompassing critical material concepts, structural characteristics, and synthesis protocols. This ensured strict semantic alignment between the source corpus and the generated queries.

**Reasoning-driven knowledge distillation:** To secure high-fidelity responses, the generated questions and corresponding source texts were processed by DeepSeek-R1. Leveraging its superior logical reasoning capabilities, the model synthesized scientifically rigorous and logically coherent answers, thereby establishing the initial instruction-response pairs.

**Human-machine collaborative verification:** A dual-layer validation protocol was implemented to guarantee scientific rigor. First, domain experts conducted random sampling evaluations to assess logical consistency and factual accuracy, iteratively refining the prompt engineering based on this expert feedback. Second, an automated self-evaluation mechanism was deployed, where DeepSeek-R1 scored each data entry (instruction, context, and response) on a scale of 0 to 5 for factuality and consistency. Consequently, only high-quality samples exceeding a threshold of 4 were retained, while substandard entries were discarded. Collectively, this pipeline yielded the final Mat-252K-SFT dataset, comprising 252,000 high-quality instruction-tuning pairs. The dataset was stratified into a 250,000-sample training set and a 2,000-sample held-out test set, providing a robust foundation for training and evaluating specialized material science models.

**4.2 MAT-R1 model**

**4.2.1 Base model selection**

To identify the optimal foundational architecture for specialized materials science applications, we conducted a systematic comparative analysis of initial language modeling (LM) loss across a spectrum of LLMs. To ensure a rigorous and fair comparison, all models were evaluated under a strictly unified configuration: a fixed random seed of 42, a consistent sample packing strategy with a maximum sequence length of 10,240 tokens, and a uniform global batch size of 16. Loss metrics were computed over the global batch. To preclude any bias arising from data stochasticity, formatting discrepancies, or batch variations, the input corpus was standardized across all candidate architectures. Furthermore, all evaluations were executed within an identical hardware and software environment to guarantee reproducibility. Complementing these internal loss metrics, we incorporated external performance benchmarks on scientific reasoning (GPQA and HLE), sourced from the Artificial Analysis leaderboard, to serve as proxies for general scientific competence.

**4.2.2 Implementation details of Mat-R1**

**Training infrastructure and data.** Mat-R1 was developed *via* full-parameter Supervised Fine-Tuning (SFT) on a composite corpus comprising the custom-curated Mat-252K-SFT dataset and the scientific reasoning subset from open-r1/Mixture-of-Thoughts (containing 173,000 multi-disciplinary trajectories). Training was implemented using the Megatron-Swift distributed framework with BF16 mixed precision, executed on a single high-performance node equipped with 8×NVIDIA H800 (80GB) GPUs.[59]

**Parallelism and optimization.** To optimize hardware efficiency for the Mixture-of-Experts (MoE) architecture, we employed a hybrid parallelism strategy, configuring Tensor Parallelism (TP) at 2 and Expert Parallelism (EP) at 8. An auxiliary loss coefficient of $10^{-3}$ was applied to enforce load balancing among experts. Computational throughput was enhanced *via* Flash-Attention 2,[60] while full activation checkpointing was utilized to accommodate extended context windows within GPU memory constraints.

**Hyperparameter configuration.** Training proceeded with a global batch size of 16 and a maximum sequence length of 16,384 tokens, utilizing sample packing to mitigate padding overhead. Optimization was performed using AdamW with a cosine decay schedule: the learning rate was linearly warmed up from $10^{-6}$ to a peak of $2\times10^{-5}$ over the initial 5% of steps, subsequently decaying to $10^{-6}$.[61] The model was trained for 2 epochs. To facilitate reproducibility, all source code, environment specifications, and configuration files are publicly available in our GitHub repository.

**4.3 Mat-T1 model**

**4.3.1 Construction of the Mat-MCP platform**

To bridge the gap between abstract reasoning and precise execution, we established Mat-MCP, a standardized tool ecosystem adhering to the Model Context Protocol (MCP). This platform serves as the instrumental foundation for Mat-T1, integrating a comprehensive suite of modules stratified into general and domain-specific categories.

**General information retrieval:** To support broad knowledge acquisition and literature mining, we integrated distinct search capabilities. This includes a real-time web retrieval tool powered by the Bing Search API for general inquiries, alongside a specialized scientific literature mining tool utilizing SearXNG to extract academic insights from distributed sources.

**Specialized materials science toolkit:** The core research capabilities are encapsulated in five functional clusters. (1) Data acquisition: We wrapped APIs for the Materials Project and OQMD to enable authoritative retrieval of crystallographic and property data. (2) Generative design: We integrated CrystaLLM for conditional lattice generation and MatterGen for diffusion-based material discovery. (3) Structural validation: Leveraging Pymatgen and the PhaseDiagram tool, we implemented rigorous protocols for atomic geometry validation, thermodynamic stability analysis, and stoichiometric plausibility assessment. (4) Property prediction: Advanced calculation modules were built upon MatGL, incorporating M3GNet, MEGNet, and CHGNet to facilitate structure relaxation, bandgap estimation, and molecular dynamics simulations. (5) Multiphysics simulation: Complementary tools include MatterSim for property prediction, FairChem for structural optimization, and VASP for high-fidelity first-principles calculations Additionally, a local knowledge retrieval tool was deployed for precise extraction from internal literature repositories.

**Infrastructure and deployment:** To ensure scalability and reproducibility, the entire toolkit was containerized *via* Docker and orchestrated on a Kubernetes (K8s) cluster. This architecture utilizes auto-scaling capabilities to guarantee robust service availability during high-throughput inference. The complete source code and protocol definitions are publicly available in our GitHub repository to foster transparency.

### 4.3.2 Implementation details of Mat-T1

**RL framework and optimization:** To endow Mat-T1 with autonomous agency for scientific exploration, we established a closed-loop Reinforcement Learning (RL) pipeline atop the Mat-MCP platform. Mat-T1, initialized from the Qwen3-14B base model, was optimized utilizing the Decoupled Clip and Dynamic sAmpling Policy Optimization (DAPO) algorithm within the VeRL distributed training framework.[62] Addressing the significant computational overhead imposed by multi-turn tool interactions, we implemented a suite of optimizations to maximize rollout throughput and memory efficiency. Specifically, we leveraged the vLLM inference engine configured with Tensor Parallelism (TP=4) to accelerate the generation of long-horizon reasoning trajectories.[63] Subsequently, policy gradients were computed using PyTorch Fully Sharded Data Parallel (FSDP), integrated with Ulysses Sequence Parallelism (SP=4) and CPU Offloading. This hybrid parallelism architecture successfully accommodates context windows of up to 16,000 tokens. To ensure robust advantage estimation, the rollout group size was set to G=8, generating eight parallel trajectories per prompt.

**Hyperparameters setting for training stability:** To ensure training stability, we adopted an asymmetric dual-clip strategy following the DAPO algorithm, with the clipping interval set to [0.2, 0.28]. Notably, we excluded the KL divergence penalty (KL coefficient $\beta = 0.0$) to encourage broad exploration of the policy space. The model was fine-tuned for 1 epoch on 8× NVIDIA H800 GPUs using the AdamW optimizer with a fixed learning rate of $1 \times 10^{-6}$ and a global batch size of 8.

**Dataset and task complexity:** The training dataset, Mat-20K-RL, is a curated subset of the SFT corpus, comprising 20,000 high-complexity queries that span structure generation, property prediction, and synthesis planning. Crucially, the task design inherently precludes "shortcut learning": the agent cannot maximize rewards by simply retrieving a theoretical value *via* a single, trivial API lookup. Instead, it is compelled to synergistically orchestrate multiple tools—generating lattices *via* CrystaLLM, relaxing structures *via* MatGL, and validating stability via molecular dynamics (MD) simulations—to maximize the acquisition of deterministic, high-value scientific results.

**Training workflow:** For a given scientific query, Mat-T1 autonomously plans and generates multiple parallel internal Chains of Thought (CoT) containing executable tool commands. These tool invocations are parsed by the vLLM engine and executed by the MCP server, and the resulting feedback (Observations) is fed back into the context window to serve as the input for the next turn. Guided by reward signals, the agent dynamically refines its strategy, iterating through a multi-turn "Reason-Act-Observe" loop until the maximum interaction limit is reached.

### 4.3.3 Design of reward function

To cultivate robust multi-turn reasoning and autonomous tool orchestration in Mat-T1, we engineered a quad-dimensional composite reward mechanism tailored for the DAPO framework. This optimization strategy aligns the policy of Mat-T1 with the iterative scientific inquiry paradigm of "Think-Plan-Execute-Observe". The total reward function, $R_{total}$, is formulated as a weighted linear combination of four distinct objective signals:

$$R_{total} = w_1 R_{turns} + w_2 R_{think} + w_3 R_{format} + w_4 R_{syntax}$$

where the hyperparameter weights are configured as $w = \{0.1, 0.3, 0.25, 0.35\}$. The rationale and formulation for each reward dimension are detailed below:

**Multi-step reasoning reward:** To mitigate the propensity of Mat-T1 for premature conclusion and foster the decomposition of complex queries into iterative subtasks, we implemented a tiered reward function contingent on the number of tool interaction turns, $n$. This mechanism assigns non-linearly increasing incentives to sustained interactions, granting the maximum reward when the trajectory depth reaches a threshold $k$. In this study, we configured $k = 4$ to effectively enforce deep investigative behavior within computational constraints:

$$R_{turns}(n) = \begin{cases} 0.0, n = 0 \\ 0.5, n = \lceil k/4 \rceil \\ 0.7, n = \lceil k/2 \rceil \\ 1.0, n \geq k \end{cases}$$

**Depth of thought reward:** Deep reasoning is a prerequisite for ensuring the correctness of tool-calling logic. In contrast to rigid, threshold-based mechanisms, we implemented a hyperbolic tangent function to model a continuous, smooth reward curve. This mechanism evaluates the average token length $\overline{L}$ of the content enclosed within <think> tags across all reasoning steps. We set the scaling factor $\lambda = 500$ to encourage the model to generate detailed and logically rigorous thought chains rather than brief intuitive judgments:

$$R_{think} = tanh(\frac{\overline{L}}{\lambda})$$

This design allows the reward score to increase smoothly with the length of reasoning, approaching approximately 0.76 when $\overline{L} \approx 500$ tokens and gradually converging to 1.0.

**Structured format reward:** To standardize the output topology of Mat-T1, we enforce strict adherence to a "Think-then-Act" paradigm. The reward function segments the generation trajectory into intermediate reasoning blocks and final response blocks, evaluating them *via* separate criteria:

$$R_{format} = \alpha \cdot \frac{1}{K} \sum_{i=1}^{K} \mathbb{I}(Think_i \prec Tool_i) + \beta \cdot \mathbb{I}(Think_{final} \prec Answer_{final})$$

We assigned a weight of $\alpha = 0.4$ to intermediate steps, mandating the presence of both <think> and <tool_call> tokens, with the critical constraint that the reasoning trace must strictly precede tool invocation ($Think \prec Tool$). This design penalizes the premature generation of answers during intermediate phases. Meanwhile, we assigned $\beta = 0.6$ to the terminal phase, requiring the co-occurrence of <think> and <answer> tags to ensure that the final conclusion is derived from explicit reasoning. Collectively, this mechanism mitigates the tendency of model to bypass the reasoning process, thereby reinforcing the structural integrity of the agentic workflow.

**Syntax reward:** To guarantee the executability of generated actions, we established a rigorous validity constraint. This mechanism subjects every tool invocation to a dual-layer verification protocol: (1) Registry Verification, which confirms the existence of the invoked function within the Mat-MCP namespace; and (2) Parameter Integrity, which validates argument completeness and type adherence *via* Pydantic schemas (e.g., ensuring a generated CIF string is crystallographically parseable by Pymatgen). The reward formulation quantifies the mean success rate of these invocations:

$$R_{syntax} = \frac{1}{M} \sum_{j=1}^{M} \mathbb{I}\left(Valid(C_j)\right)$$

Here, $M$ denotes the total number of tool calls in a trajectory, and $\mathbb{I}$ represents the indicator function. Crucially, a positive reward is contingent upon the call being both syntactically well-formed (JSON-parsable) and semantically valid (execution-ready parameters). This constraint effectively bridges the gap between abstract decision-making and concrete execution, ensuring that all planned actions are viable within the Mat-MCP infrastructure.

**4.4 Implementation of MatBrain**

We architected the MatBrain system as a graph-based state machine leveraging LangGraph. This framework employs a Directed Cyclic Graph (DCG) to orchestrate the asynchronous, collaborative workflow between the Mat-T1 and Mat-R1 models. As depicted in Figure 1b, the workflow is formally defined by a network of nodes and edges that transmit context, execution results, and task metadata *via* a shared global state. The complete implementation code and deployment scripts are publicly available in our GitHub repository to facilitate reproducibility. The system implementation is structured into four key components:

**Execution Node (Mat-T1):** The workflow initiates at the execution node, driven by the Mat-T1 model. This component is responsible for translating user queries or feedback from previous turns into concrete tool planning and execution actions. To guarantee robustness, we embedded a Runtime Parameter Validation Layer within this node. Utilizing Pydantic schemas, the system performs real-time type checking and format verification on all JSON parameters generated by Mat-T1 (e.g., rigorously validating the syntactic compliance of generated CIF crystal files). Only validated instructions are dispatched to the Mat-MCP server. Upon completion, the node captures standard output (stdout) and error streams (stderr), encapsulating them into standardized observations to update the global state.

**Reasoning node (Mat-R1):** Data flows from execution to the reasoning node, which hosts Mat-R1. Distinct from monolithic ReAct loops, Mat-R1 operates as a pure analytical engine decoupled from the tool environment. It ingests the cumulative execution history from the global state, leveraging its reasoning capabilities to perform physical plausibility assessments on the complex data returned by materials science tools. The output of this node is structured into two distinct components: a scientific interpretation of the current experimental results and a strategic intent for the subsequent action.

**Dynamic routing logic:** The control flow of the system is governed by a conditional edge mechanism (represented by the decision diamond in Figure 1b). We established a dynamic routing protocol by parsing Mat-R1's decision intent: (1) Iterative Rollback: If Mat-R1 deems the current evidence insufficient (e.g., structure optimization non-convergence or missing band structure data), the system extracts the generated "next instruction", reverts control to the Mat-T1 node, and triggers a new execution cycle. (2) Termination: if Mat-R1 concludes the task is resolved, the route directs the flow to a termination state to extract the final scientific answer.

**Concurrency and safety:** To prevent infinite recursion and ensure computational efficiency, a maximum recursion depth (max_iterations, default=6) is enforced within the graph structure. Upon reaching this threshold, the system compels Mat-R1 to generate a best-effort prediction based on the current state. The entire architecture is built upon asyncio of the Python framework, ensuring efficient, non-blocking parallel handling of dual-model inference and I/O-intensive tool interactions.

**4.5 Evaluation of NRR performance**

**Preparation of the working electrode:** First, a piece of CP (1×1 cm$^2$) was cleaned *via* sonication with ethanol and water several times and dried in the oven. Then, 10 mg of $CoV_4S_8$ nanosheets and 20 μL of Nafion solution (5 wt% %) were dispersed in 980 μL ethanol, followed by 10 min sonication to form a homogeneous black dispersion. Finally, 100 μL of ink was dropped on CP and dried under ambient conditions.

**Electrochemical measurements:** The electrochemical NRR performance measurements were carried out in a gastight H-type cell separated by Nafion 117 membrane using a CHI 760E electrochemical station (Chenhua, Shanghai). The electrochemical experiments were performed in a three-electrode system, with the graphite rod as the counter electrode, SCE as the reference electrode. All experiments were performed at ambient conditions. Before the test, the 0.1 M HCl electrolyte in the cathode chamber was bubbled with high-purity $N_2$ or Ar (99.999%) for 30 min at a flow rate of 30 mL min$^{-1}$. All potentials reported in this work were converted to reversible hydrogen electrode (RHE) scale *via* the equation: $E_{RHE} = E_{SCE} + 0.059 \times pH + 0.242$.

**Determination of NH$_3$:** The Indophenol blue method was used to quantify the produced NH$_3$ in electrolyte by UV-Vis spectrophotometer. 2 mL of 1 M NaOH solution containing 5 wt% salicylic acid and 5 wt% sodium citrate, 1 mL of 0.05 M NaClO, and 0.2 mL of 1 wt% sodium nitroferricyanide dihydrate were added into 2 mL of electrolyte in sequence. After keeping in dark for 2 h, the UV-Vis absorption was measured at a wavelength range of 800 ~ 500 nm.

**Determination of N$_2$H$_4$:** The Watt and Chrisp method was applied to detect N$_2$H$_4$ in the HCl electrolyte. The mixture solution of 1.0 g *p*-C$_9$H$_{11}$NO, 5 mL HCl, and 100 mL ethanol was used as a color reagent. 2 mL of prepared color reagent was added into 2 mL electrolyte. After keeping in dark for 15 min, the UV-Vis absorption was measured at a wavelength range of 500 ~ 400 nm.

**Calculation of NH$_3$ yield and FE:** NH$_3$ yield was calculated using the equation:

$$NH_3 \text{ yield} = C_{NH3} \times V / (m_{NH3} \times t)$$

Where $C_{NH3}$ is the concentration of NH$_3$ in catholyte, V is the volume of catholyte (30 mL), $m_{NH3}$ is the loaded mass of catalyst on carbon paper (0.1 mg), and *t* is the time for which the potential was applied (2 h).

FE was calculated according to the equation:

$$FE = 3 \times F \times C_{NH3} \times V / (17 \times Q)$$

Where 3 is the number of electrons to produce one NH$_3$ molecule, F is the Faraday constant (96485 C mol$^{-1}$); $C_{NH3}$ is the measured mass concentration of NH$_3$; V is the volume of catholyte (30 mL); 17 is the relative molecular mass of NH$_3$; Q is the quantity of applied charge.

## Data availability

The data supporting the findings of this study are available within the paper and in the supplemental information files. Source data can be found at https://github.com/MAIC-SIAT/matbrain.

## Code availability

In our commitment to transparency and reproducibility, we have released our code showing our implementation. Python scripts for MatBrain developed in this study are available at https://github.com/MAIC-SIAT/matbrain.

## Acknowledgements

This work was financially supported by National Key R&D Program of China (2023YFA0915600), the Shenzhen Medical Research Fund (D250402005), National Natural Science Foundation of China (32471459), Strategic Priority Research Program of the Chinese Academy of Sciences (XDB0930000), Shenzhen Science and Technology Program Grant (RCJC20200714114435061, KJZD20230923114703007), Guangdong Basic and Applied Basic Research Foundation (2025B1515020088), Natural Science Foundation of Hunan Province (2025WK2013), Guangdong Provincial Key Laboratory of Multimodality Non-Invasive Brain-Computer Interfaces (2024B1212010010), Original Innovation Project in SIAT (JQ0209-2025), the China Postdoctoral Science Foundation (2025M780875), the Postdoctoral Fellowship Program of CPSF (GZC20241848), Sichuan Science and Technology Program (2025ZNSFSC0900).

## Author contributions

T.S. and Y.L. contributed to the conception and design of the work. Z.L. and Q.L. contributed to the experiment execution and technical implementation. T.S., Y.L. and J.W. provided domain expertise for the literature review and validation of rules. J.Z., W.X. and L.Y. contributed to the experimental methodology. D.D. provided hardware support. T.S. prepared the figures. T.S., Y.L., Z.L., Q.L. and X.Y. wrote the draft manuscript. R.H., W.Z., and J.W. contributed to the revision of the manuscript. X.Y. supervised the project.

## Competing interests

The authors declare no competing interests.